%% file: AnonymousSubmission2027.tex
\documentclass[letterpaper]{article} % DO NOT CHANGE THIS
\usepackage{aaai2027}  % DO NOT CHANGE THIS
\nocopyright % arXiv preprint: no AAAI publication copyright claim
\usepackage[hyphens]{url}  % DO NOT CHANGE THIS
\usepackage{graphicx} % DO NOT CHANGE THIS
\usepackage{natbib}  % DO NOT CHANGE THIS AND DO NOT ADD ANY OPTIONS TO IT
\usepackage{caption} % DO NOT CHANGE THIS AND DO NOT ADD ANY OPTIONS TO IT
\usepackage{algorithm}
\usepackage{algorithmic}

\usepackage{amsmath}
\usepackage{amssymb}
\usepackage{xcolor}

\usepackage{newfloat}
\usepackage{listings}
\DeclareCaptionStyle{ruled}{labelfont=normalfont,labelsep=colon,strut=off} % DO NOT CHANGE THIS
\floatstyle{ruled}
\newfloat{listing}{tb}{lst}{}
\floatname{listing}{Listing}

\usepackage{booktabs}
\usepackage{tabularx}
\usepackage{amsmath}
\usepackage{amssymb}
\usepackage{subfigure}
\usepackage{multirow}
\usepackage{colortbl}

\usepackage[most]{tcolorbox}

\newtcolorbox{promptbox}[2]{
  breakable,
  colback=white,
  colframe=black,
  coltitle=black,
  colbacktitle=gray!20,
  fonttitle=\bfseries,
  title={#1},
  title after break={#2\ \textit{(continued)}},
  boxrule=0.6pt,
  arc=0pt
}

\newtcolorbox{promptfigurebox}[1]{
  colback=white,
  colframe=black,
  coltitle=black,
  colbacktitle=gray!15,
  fonttitle=\bfseries,
  title={#1},
  boxrule=0.6pt,
  arc=0pt
}

\newtcolorbox{casebox}[1]{
  colback=white,
  colframe=black,
  coltitle=white,
  colbacktitle=black,
  fonttitle=\bfseries,
  title={#1},
  boxrule=0.6pt,
  arc=0pt
}

\title{CuratorMAS: Automating Dataset Curation via Multi-Agent Orchestration}
\author {
    Yixin Zhang\textsuperscript{\rm 1},
    Wenjie Feng\textsuperscript{\rm 1}\corresponding
}
\affiliations {
    \textsuperscript{\rm 1}University of Science and Technology of China\\
    yixinzhang@mail.ustc.edu.cn, fengwenjie@ustc.edu.cn
}

\begin{document}

\maketitle

\begin{abstract}
High-quality datasets are essential for reliable machine learning, but dataset curation remains 
costly and hard to generalize across domains. Existing methods typically rely on manually designed heuristics or 
model-dependent signals, limiting their applicability across tasks and user queries. 
To address these limitations and automate data curation, we propose \textbf{CuratorMAS}, 
a multi-agent collaboration framework that orchestrates agents to evaluate and curate high-quality datasets.  
To achieve the goal of flexible curation, CuratorMAS decomposes the complex curation process into 
five programmable execution stages and forms a parallelizable workflow. Specifically, CuratorMAS 
first performs dataset exploration to collect contextual information such as file structures and 
constraint cues, thereby developing a comprehensive understanding of the given task. 
In order to acquire up-to-date information, CuratorMAS retrieves domain knowledge from online sources to 
augment the evaluation process. Next, CuratorMAS derives the necessary evaluation criteria 
and computes the corresponding metrics. Based on these results, CuratorMAS executes filtering accordingly. 
Finally, an evolution module summarizes the evaluation outcomes and updates the relevant skills. 
Extensive and comprehensive experiments demonstrate that CuratorMAS significantly 
reduces the noise rate by up to 36.03 percentage points (pp) while also improving the F1 score of  
downstream models by up to 8.88 pp.

\end{abstract}

    % \link{Code}{https://anonymous.4open.science/r/CuratorMAS}
    % \link{Datasets}{https://aaai.org/example/datasets}
    % \link{Extended version}{https://aaai.org/example/extended-version}

\section{Introduction}

High-quality data are fundamental to the success of modern machine learning systems, 
but dataset curation remains costly, domain-specific, and difficult to scale~\citep{FineWeb,Dolma}. As training datasets 
are increasingly collected from web sources, crowdsourced annotations, and synthetically generated samples, 
dataset curation faces three key challenges: (i) how to automatically construct task-specific datasets; 
(ii) how to ensure the quality of the curated datasets; and (iii) how to satisfy diverse user requirements. 
Failure to address these challenges can hinder downstream model training by slowing convergence, 
amplifying annotation bias, and degrading generalization performance.

Large-scale data curation pipelines typically rely on predefined heuristic rules, such as 
text-length thresholds and toxicity filters~\citep{FineWeb,Dolma,C4}. Although  
computationally efficient, these heuristics are often brittle and inflexible when the task or 
user query changes. Another line of work estimates data quality via model-dependent signals, 
including training dynamics, confident learning, and meta-learned data valuation~\citep{ConfidentLearning,DivideMix,ProMix,SelfMix,LossRank,VUsableInformation,DataRater}. 
While these methods provide useful estimates of data quality, they often require task-specific model training 
or access to model internals, limiting their applicability to diverse user-provided datasets.  
Recent advances in large language models (LLMs) have demonstrated strong capabilities in reasoning, 
tool use, and data analysis~\citep{LLMsStrugglewithDataAnalysis,DataAnalyticAgents}, making LLM-based evaluation a promising alternative~\citep{zhang2023llmaaa,ACTasHuman}.
Nevertheless, a central question remains: \textit{How can agents automatically curate and evaluate datasets across diverse domains and user requirements?}
 
\begin{figure*}[t]
  \centering
  \includegraphics[width=0.93\textwidth]{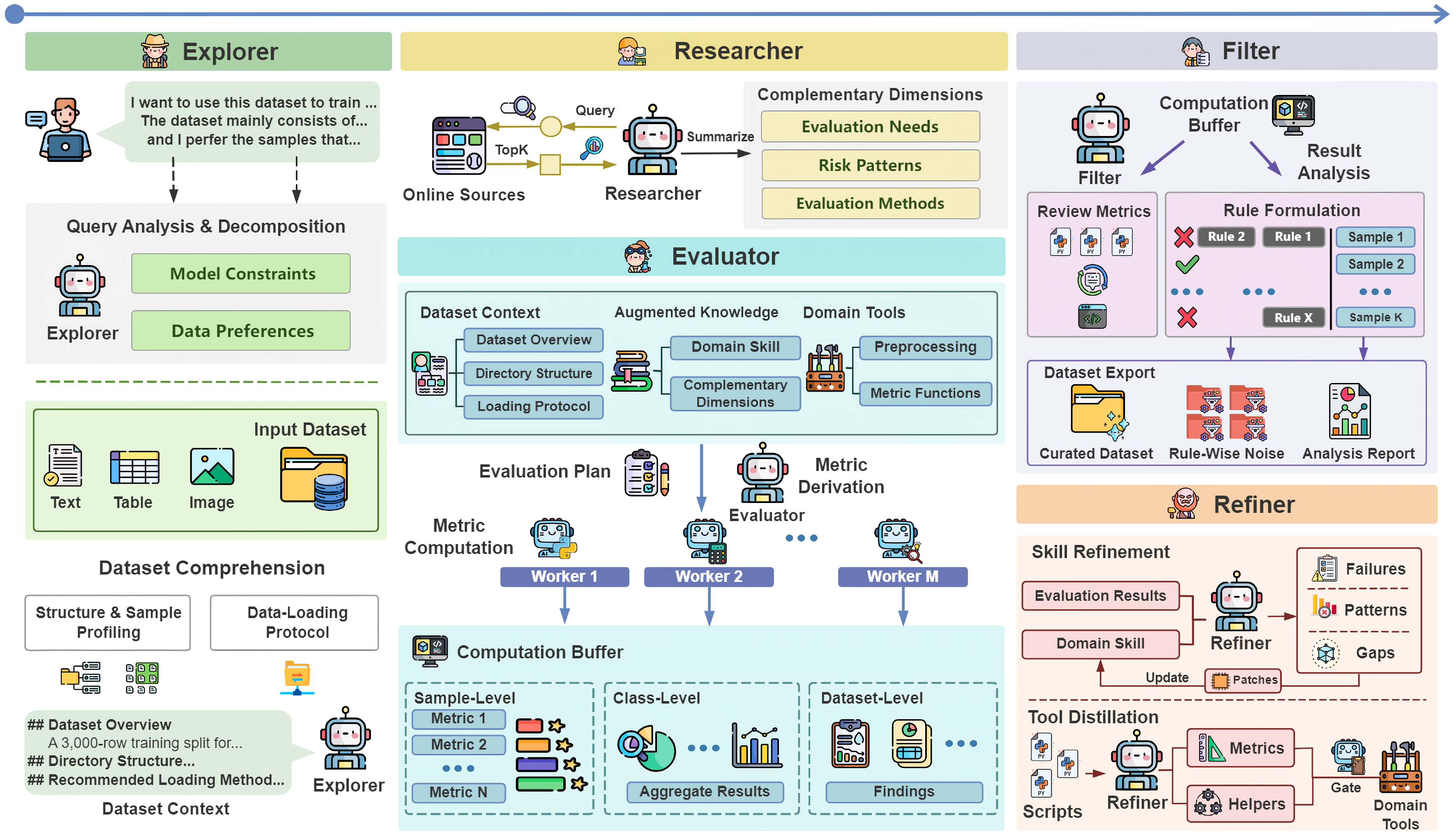}
  \caption{Overview of the CuratorMAS framework. }
  % \caption{Overview of the CuratorMAS framework. Specifically, upon receiving the input dataset and 
  % a user query, CuratorMAS summarizes the dataset and translates the query into evaluation constraints. It 
  % then retrieves external knowledge to enrich the domain skills, thereby supporting the evaluation. 
  % Based on the evaluation results, CuratorMAS derives and applies appropriate filtering criteria to 
  % produce a curated dataset and updates the relevant domain skills. }
  \label{fig:methodology}
\end{figure*}

To this end, we propose \textbf{CuratorMAS}, an automated multi-agent collaboration framework that 
orchestrates four core curation agents---an Explorer Agent, a Researcher Agent, 
an Evaluator Agent, and a Filter Agent---to curate and evaluate high-quality datasets. Given a dataset and 
a user query, CuratorMAS first performs dataset comprehension to analyze the dataset structure and identify constraints imposed by downstream models and user preferences. 
It then conducts knowledge-augmented quality evaluation, during which the researcher retrieves 
up-to-date domain knowledge to enrich the system's evaluation skills, while the evaluator 
derives appropriate quality criteria and computes the corresponding 
metrics. Based on the results, the filter agent makes criterion-aware 
filtering decisions and finally produces a curated dataset for downstream 
use. Unlike static rule-based and model-dependent approaches, CuratorMAS adapts its 
evaluation criteria to different datasets while providing transparent and auditable evaluation results.
In addition, its evolution module enables the system to continuously incorporate up-to-date knowledge across runs.

Our primary contributions are as follows: (i) We explicitly break down the data curation process into an executable and 
programmable workflow, providing a structured paradigm for automated data curation. (ii) We present CuratorMAS, a 
multi-agent collaboration framework that coordinates agents to flexibly evaluate and curate datasets across different 
domains. (iii) We evaluate CuratorMAS against six baselines on two datasets spanning different modalities and domains, 
including rule-based, model-based, and prompt-based methods. CuratorMAS reduces the noise rate by up to 36.03 pp, 
while improving the F1 scores of the downstream models by up to 8.9 pp through our effective workflow.

\section{Related Work}

\paragraph{Dataset Curation.}
Large-scale dataset construction typically relies on computationally efficient heuristics, such as length and toxicity 
filtering~\citep{FineWeb,Dolma,C4}, but these rules are dataset-specific and require manual redesign across modalities, 
tasks, and user needs. Model-based techniques, including confident learning, influence-based valuation, and meta-learned 
utility estimation~\citep{DatasetCartography,ConfidentLearning,VUsableInformation,DataRater}, provide quality 
estimates but are often tied to specific task domains and are difficult to generalize. Building on recent work using 
multimodal LLMs for quality assessment~\citep{ACTasHuman}, our framework employs LLMs for query- and domain-aware 
dataset evaluation and curation.

\paragraph{Multi-Agent Collaboration.}
Prior studies on multi-agent systems have demonstrated the benefits of role specialization, debate, and workflow decomposition. 
Role-playing frameworks assign agents distinct responsibilities for collaborative reasoning~\citep{CAMEL,GDesigner}, 
while debate-based methods can improve factuality by enabling agents to critique one another~\citep{MultiAgentDebate}. 
Engineering-oriented frameworks such as AutoGen~\citep{AutoGen}, MetaGPT~\citep{MetaGPT}, and ChatDev~\citep{ChatDev} 
support the decomposition of complex tasks into workflows. CuratorMAS extends these principles to dataset 
curation by decomposing the curation process into specialized stages coordinated through a structured multi-agent workflow.

\paragraph{Self-Evolution.}
Agent self-evolution enables agents to improve by transforming experience into persistent memories, reusable skills, or executable tools. 
Reflexion~\citep{Reflexion} stores linguistic feedback in episodic memory, while ExpeL~\citep{ExpeL} distills 
reusable insights from past trajectories. Recent studies have explored skill evolution, 
long-term memory, and the automated construction of reusable tools~\citep{Alita,SkillRL,AlitaG}. 
Unlike prior work on general reasoning and programming, CuratorMAS applies self-evolution to 
dataset curation by refining domain skills and developing reusable tools.

\section{CuratorMAS}

In this section, we present CuratorMAS, a multi-agent collaboration framework for  
curating and evaluating high-quality datasets, as illustrated in Figure~\ref{fig:methodology}. 
CuratorMAS organizes its five-stage workflow into three core components. 
1) \textit{Dataset Comprehension}: Given an input dataset and a user query, the Explorer Agent 
analyzes the dataset structure, summarizes its key characteristics, and decomposes the 
user query into constraints such as model requirements and user preferences. 
2) \textit{Knowledge-Augmented Evaluation}: In this open-domain setting, the Researcher Agent identifies gaps in the 
evaluation criteria of the current domain skill and retrieves up-to-date domain 
knowledge from online sources to augment the evaluation. Subsequently, the Evaluator 
Agent performs multi-granularity evaluation, after which the Filter Agent filters noisy 
samples accordingly to produce a curated dataset. 
3) \textit{Experience-Based Self-Evolution}: Based on the evaluation results, the Refiner Agent summarizes 
the evaluation outcomes and updates the domain skills by refining the criteria and distilling usable 
scripts. The overall workflow of CuratorMAS is formalized in Algorithm~\ref{alg:autocuration}.

\subsection{Dataset Comprehension}

To handle dataset heterogeneity across modalities and domains, we first employ an 
Explorer Agent $\pi_{\mathrm{explorer}}$ to develop a comprehensive understanding of the target dataset $D$. 
The Explorer Agent analyzes the dataset's structure, parses data files, and performs essential preprocessing. 
It then summarizes the dataset into a structured document $\mathcal{U}_{D}$, which contains a structural description $S_{D}$, 
a sample overview $O_{D}$, and a data-loading protocol $L_{D}$.  
This document further provides guidance for subsequent agent collaboration.
The Explorer Agent also decomposes the user query $q$ into 
model constraints $C_{\mathrm{model}}$ and data preferences $R_{\mathrm{data}}$. 
Together, these components define the evaluation constraint set
$\mathcal{C}_{q}=\{C_{\mathrm{model}},R_{\mathrm{data}}\}$, which bounds the retrieval and evaluation procedures. 
By providing downstream agents with a shared representation of the dataset and user requirements, 
this comprehension stage promotes consistent reasoning and reduces the context overhead 
caused by repeated exploration.

\subsection{Knowledge-Augmented Evaluation}

\paragraph{External Knowledge Retrieval.} According to the Hugging Face taxonomy, datasets 
span 50 domains and more than nine modalities, each potentially requiring distinct quality criteria. 
This diversity makes it impractical to define a fixed set of universally applicable evaluation metrics in advance.
Therefore, we introduce an online knowledge retrieval stage to support adaptive evaluation. 
Specifically, guided by the query-specific constraints $\mathcal{C}_{q}$, the Researcher 
Agent $\pi_{\mathrm{researcher}}$ reviews the current domain skill $S_{\mathrm{domain}}$ 
and identifies potential quality risks that are not yet captured. It then formulates 
$N$ search queries $\{x_i\}_{i=1}^N$ for online information retrieval. 
For each query, the Researcher Agent retrieves the top-$K$ most relevant documents $\{d_{i,j}\}_{j=1}^K$ and spawns  
subagents to summarize them as defined in Equation~\eqref{eq:retrieval_process}:
\begin{equation}
\begin{aligned}
\{s_{i,j}\}_{j=1}^{K}
&\sim
\pi_{\mathrm{subagent}}
\left(\cdot \mid \{d_{i,j}\}_{j=1}^{K}\right)
\end{aligned}
\label{eq:retrieval_process}
\end{equation}
where $\pi_{\mathrm{subagent}}$ denotes the summarization policy used by each spawned subagent, 
and $s_{i,j}$ denotes the summary of the retrieved document $d_{i,j}$. We set $N=4$ and $K=5$ to balance retrieval quality and efficiency. 
Finally, the Researcher Agent synthesizes these summaries and derives a set of
complementary evaluation dimensions $Q_{\mathrm{comp}}$, designed to cover 
previously identified quality risks that are not adequately addressed by the existing criteria,  
as follows:
\begin{equation}
Q_{\mathrm{comp}}
\sim \pi_{\mathrm{researcher}}
\left(\cdot \mid \mathcal{U}_{D}, \{s_{i,j}\}_{j=1}^{K}, S_{\mathrm{domain}}\right).
\label{eq:comp_dimensions}
\end{equation}
This stage enables CuratorMAS to adapt to diverse domains 
by incorporating up-to-date domain knowledge, thereby enhancing its 
adaptability and evaluation robustness.

\input{algorithm.tex}

\paragraph{Knowledge-Augmented Evaluation.}
The Evaluator Agent $\pi_{\mathrm{evaluator}}$ constructs an augmented evaluation knowledge base
$\mathcal{K}_{\mathrm{aug}} = S_{\mathrm{domain}} \oplus Q_{\mathrm{comp}}$, where $\oplus$ denotes a knowledge-integration operator.
Based on $\mathcal{K}_{\mathrm{aug}}$, the agent performs knowledge-augmented evaluation.  
Specifically, we adopt a two-step process consisting of  
metric derivation and metric computation, thereby decoupling high-level metric design from error-prone computation.

During metric derivation, the Evaluator Agent leverages the dataset context 
$\mathcal{U}_{D}$, the augmented knowledge $\mathcal{K}_{\mathrm{aug}}$, and 
the task constraints $\mathcal{C}_{q}$ to operationalize high-level evaluation dimensions as 
concrete, computable metrics, as illustrated in Equation~\eqref{eq:knowledge_augmented_planning}. 
It then organizes the derived metrics into a structured evaluation plan $P_{\mathrm{eval}}$
spanning three granularities: a sample-level component $P_{\mathrm{sample}}$, a class- or group-level component 
$P_{\mathrm{class}}$, and a dataset-level component $P_{\mathrm{dataset}}$. 
\begin{equation}
\begin{aligned}
P_{\mathrm{eval}}
&\sim
\pi_{\mathrm{evaluator}}
\left(
\cdot \mid \mathcal{U}_{D}, \mathcal{K}_{\mathrm{aug}}, \mathcal{C}_q
\right)
\end{aligned}
\label{eq:knowledge_augmented_planning}
\end{equation}
During the computation, we dispatch multiple Worker Agents to execute evaluation tasks in parallel to ensure 
computational efficiency. Each Worker Agent operates in an isolated workspace with a dedicated
Python interpreter, thus preventing cross-task context interference. Upon receiving an assignment $t_m \in P_{\mathrm{eval}}$, 
a Worker Agent loads the required files, performs the computation, and writes the results to the shared 
computation buffer $\mathcal{B}$, as formalized in Equation~\eqref{eq:parallel_evaluation}.
\begin{equation}
\mathcal{B}
=
\operatorname{Buffer}\!\left(
\left\{
o_m \,\middle|\,
o_m \sim \pi_{\mathrm{worker}}(\cdot \mid t_m,\mathcal{U}_D)
\right\}_{m=1}^{M}
\right)
\label{eq:parallel_evaluation}
\end{equation}
where $M=|P_{\mathrm{eval}}|$ is the number of metrics in the evaluation plan and $o_m$ denotes 
the output produced by task $t_m$.

\paragraph{Agentic Filtering and Dataset Export.} 
Using the results stored in the computation buffer $\mathcal{B}$, the Filter Agent $\pi_{\mathrm{filter}}$ performs a 
second-stage analysis of the computed metrics and associated findings to identify 
poor-quality samples. It reviews the evaluation results and formulates a set of filtering 
rules $\{r_{\ell}\}_{\ell=1}^{L}$ based on the identified issues and the query-specific constraints $\mathcal{C}_{q}$. 
Each rule is applied independently, and the samples flagged by each rule are recorded for traceability. 
The union of the flagged samples is then removed from the original dataset. 
Finally, the remaining samples are exported as 
the curated dataset $D_{\mathrm{curated}}$ for downstream use. CuratorMAS also produces a detailed 
evaluation report and a set of visualizations summarizing the evaluation results. The overall 
filtering process is formalized in Equation~\eqref{eq:agentic_filtering}.
\begin{equation}
\begin{aligned}
D_{\mathrm{noise}}^{(\ell)}
&= \{x\in D:r_{\ell}(x)=1\}, \\
D_{\mathrm{curated}}
&= D\setminus\bigcup_{\ell=1}^{L}D_{\mathrm{noise}}^{(\ell)}.
\end{aligned}
\label{eq:agentic_filtering}
\end{equation}
Here, each filtering rule is defined as an indicator function $r_{\ell}:D\rightarrow\{0,1\}$, where $r_{\ell}(x)=1$ 
means that sample $x$ is identified as noisy, and $r_{\ell}(x)=0$ otherwise.
Accordingly, the noisy subset identified by $r_{\ell}$ is formalized as $D_{\mathrm{noise}}^{(\ell)}$.

\subsection{Experience-Based Self-Evolution}

\paragraph{Iterative Skill Refinement.} 
As user requirements vary across domains, a static knowledge base may not adequately cover diverse  
curation scenarios and may therefore provide insufficient guidance. To tackle this issue, 
CuratorMAS iteratively refines the domain skills to support continual learning. 

The Refiner Agent $\pi_{\mathrm{refiner}}$ reviews the computed metrics and evaluation findings 
to identify capability gaps (i.e., overly general criteria or insufficiently covered cases) 
and failure modes in the current domain skill $S_{\mathrm{domain}}$. 
It then derives a set of patches $S_{\mathrm{patch}}$ to remedy these deficiencies and update the 
domain skills as follows:
\begin{equation}
    S'_{\mathrm{domain}} \leftarrow S_{\mathrm{domain}} \cup S_{\mathrm{patch}}
\end{equation}

\paragraph{Reusable Tool Distillation.} 
During evaluation, the agents may generate a substantial number of execution scripts for common data-processing operations.
To leverage these one-off snippets, the Refiner Agent distills them into compact, reusable data-processing tools. 
We maintain a domain-specific tool library $T_{\mathrm{domain}}$. 
Rather than modifying the library directly, the Refiner Agent submits proposed changes 
as a set of CRUD operations, denoted by $T_{\mathrm{patch}}$.  
Before $T_{\mathrm{patch}}$ is applied to $T_{\mathrm{domain}}$, another Gate 
Agent $\mathbb{I}_{\mathrm{gate}}$ is invoked to review the proposed patches and reject 
redundant, overly specific, or invalid updates. Only approved patches are committed to the tool library. 
We further adopt a progressive-disclosure strategy to avoid overwhelming the agent with a batch of raw code.
This gating mechanism improves the compactness, reusability, 
and general applicability of the tool library, as formalized in Equation~\eqref{eq:tool_distillation}:
\begin{equation}
\begin{aligned}
T'_{\mathrm{domain}}
&\leftarrow
T_{\mathrm{domain}}
\cup
\mathbb{I}_{\mathrm{gate}}(T_{\mathrm{patch}}).
\end{aligned}
\label{eq:tool_distillation}
\end{equation}

\input{tables/contamination_comparision.tex}

\section{Experiments}
\subsection{Experimental Settings}

\paragraph{Datasets.}
We evaluate CuratorMAS on three datasets across the image and text modalities. The text benchmarks are 
(i) JD: a sentiment classification dataset of product reviews from the JD 
platform~\cite{damo-nlp-jd}. We selected and manually re-annotated 1,000 samples to construct a clean subset, 
denoted as JD\_clean; (ii) NoisyAG-News: a human-annotated news-topic  
classification dataset with naturally occurring label noise~\cite{NoisyAGNews}. We selected 3,000 samples with a noise rate
of 19.83\%. The image benchmark is (iii) CIFAR-10N: a human-annotated image classification dataset
containing real-world label noise~\cite{CIFAR10N}. We selected 3,000 samples with a noise rate of 39.67\%.

\input{tables/baseline_real_world_comparison.tex}

\paragraph{Baselines.}
We compare CuratorMAS with six baselines across three categories: rule-based, model-based, and prompt-based approaches. 
For the rule-based method, following previous work~\citep{FineWeb,Dolma,C4}, we implement (i) Heuristic Rules, which comprises text-integrity 
checks, anomalous-image-resolution checks, and other heuristics. Model-based methods include  
(ii) CleanLab~\citep{ConfidentLearning}, which uses confident learning to detect label noise; (iii) DivideMix~\citep{DivideMix}, which models 
the distribution of training losses to distinguish clean samples from noisy ones; 
(iv) ProMix~\citep{ProMix}, which combines small-loss selection, pseudo-label expansion, and mixed-sample augmentation; and (v) LossRank~\citep{LossRank}, which leverages loss 
as an indicator. For the prompt-based method, we evaluate (vi) LLM-as-a-Judge, which uses GPT-5.4~\citep{openai2026gpt54} as a 
judge to classify whether a sample is noisy.

\paragraph{Downstream Models.}
For image classification, we consider (i) ResNet-18~\cite{ResNet18}; (ii) MobileNetV2~\cite{MobileNetV2}; (iii) 
EfficientNet-B0~\cite{EfficientNet}; and (iv) GhostNet~\cite{GhostNet}. All four models are initialized with 
weights pretrained on ImageNet-1K~\citep{ImageNet}. For text classification, we leverage four pretrained language models, including (v) BERT-base-uncased~\cite{BERT}; 
(vi) DeBERTa-v3-small~\cite{DeBERTaV3}; (vii) MobileBERT~\cite{MobileBERT}; and (viii) BERT-tiny~\cite{WellReadStudents}.

\paragraph{Implementation Details.}
For the primary results, we use GPT-5.4 by default and also evaluate GPT-5.3-Codex~\cite{openai2026gpt53codex}
and Claude Sonnet-4.6~\cite{anthropic2026claudesonnet46}. All LLM-based agents follow the ReAct reasoning-and-action paradigm~\citep{ReAct}.
All experiments were conducted on Ubuntu 22.04 with an NVIDIA RTX 4090 GPU.

\begin{figure*}[t]
  \centering
     \subfigure[Model Performance on CIFAR-10N.\label{fig:image_downstream}]{  % over different subsets
        \includegraphics[width=0.84\linewidth]{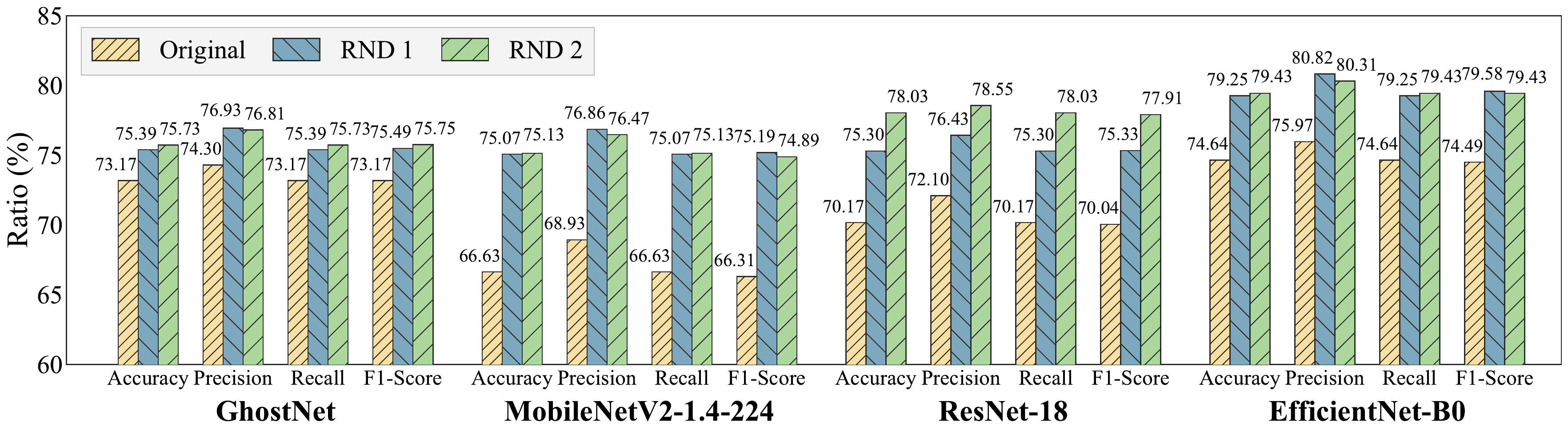}}
     \subfigure[Model Performance on NoisyAG-News\label{fig:text_downstream}]{
        \includegraphics[width=0.83\linewidth]{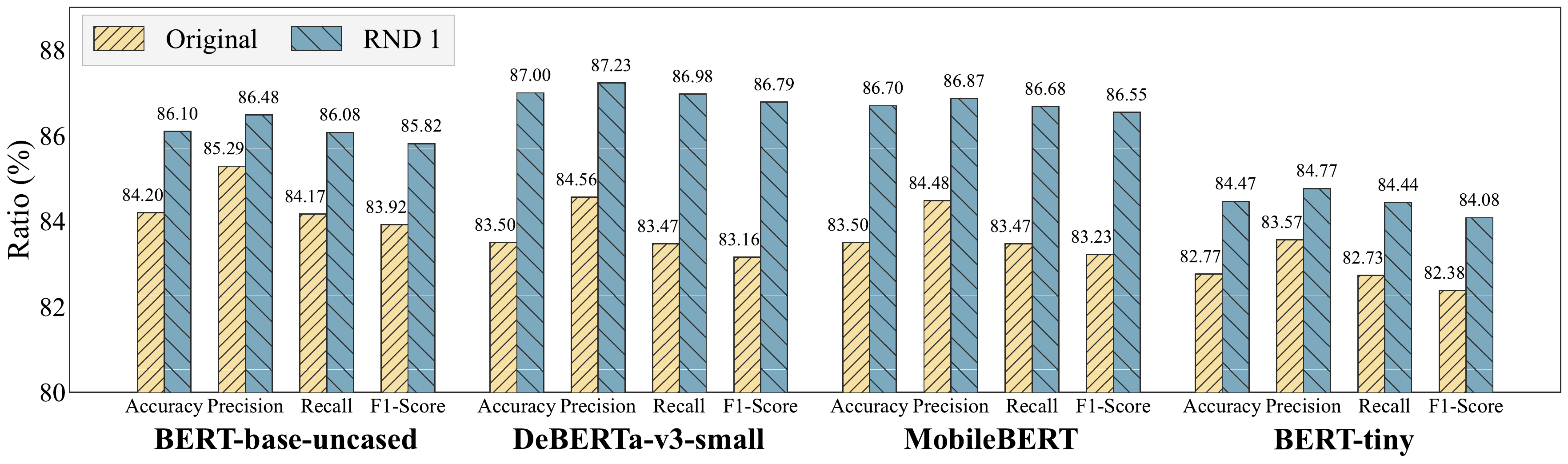}}
  \caption{Downstream performance of models trained on the original noisy and CuratorMAS-curated datasets.}
  \label{fig:downstream_performance}
\end{figure*}

% \begin{figure*}[t]
%   \centering
%   \includegraphics[width=0.87\linewidth]{Figures/image-downstream-bar.pdf}
%   \caption{Downstream performance of models trained on the original noisy
%   and CuratorMAS-curated CIFAR-10N datasets.}
%   \label{fig:image_downstream}
% \end{figure*}

\begin{figure}[t]
  \centering
  \includegraphics[width=1\linewidth]{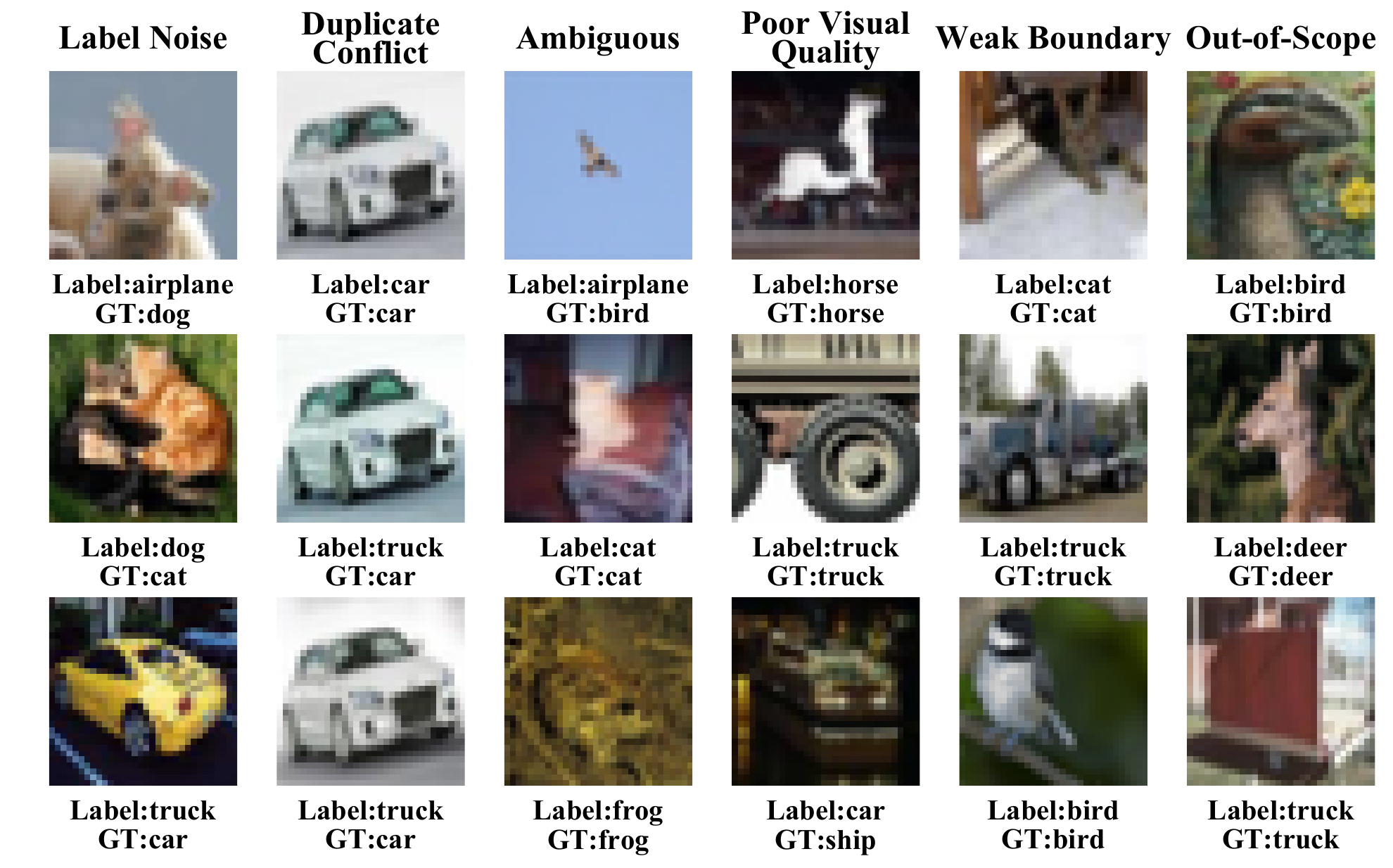}
  \caption{Examples of different noise types in CIFAR-10N. }
  % \caption{Examples of different noise types in CIFAR-10N. Each column represents
  % a specific noise category, and each image is annotated with its noisy and
  % ground-truth labels.}
  \label{fig:cifar_noise_examples}
\end{figure}

\subsection{Controlled Contamination}
\label{sec:controlled_contamination}

We synthetically inject five predefined noise types into JD\_clean at varying contamination levels. Specifically, we 
consider five representative types of noise that may arise during data collection and annotation: 
(i) \textit{Label Flipping}, where the original label is changed to the opposite class; 
(ii) \textit{Off-Domain}, where the original review is replaced with irrelevant text from an unrelated corpus, such as THUCNews~\citep{THUCTC};
(iii) \textit{Ambiguous}, where an LLM rewrites the review to make its sentiment ambiguous; 
(iv) \textit{Broken Text}, where the review is corrupted by encoding errors, HTML residues, or other textual corruption; and 
(v) \textit{Duplication}, where a randomly selected review is duplicated to replace another sample. 
These noise types are injected in equal proportions to produce overall noise rates of 10\%, 20\%, and 30\%.
For each level, we perform a single filtering pass. 

As shown in Table~\ref{tab:controlled_contamination_main}, CuratorMAS achieves favorable performance across 
noise levels while maintaining a low false-positive rate (FPR). At the 30\% level,  
CuratorMAS achieves a recall of 83.00\% and a precision of 92.72\%, with a relatively low 
FPR of 3.44\%. As the noise increases from 10\% to 30\%, CuratorMAS remains effective, with the F1 score rising from 78.22\% to 
87.13\%, whereas the FPR increases only moderately from 2.80\% to 3.44\%. This trend may be attributed to the 
increased prevalence of noisy samples, which provides more detectable positive instances. These results indicate 
that the programmable curation workflow can identify and filter multiple types of noise across noise levels.

\subsection{Real-World Label Noise}
\label{sec:real_world_scenario}

To evaluate the performance under real-world noise, we compare CuratorMAS with six classic baselines across two noisy datasets: CIFAR-10N and NoisyAG-News. We apply two rounds of filtering to CIFAR-10N and 
one round to NoisyAG-News. As shown in Table~\ref{tab:filtering_baseline_comparison}, 
CuratorMAS achieves the best overall performance among all evaluated methods. 

On CIFAR-10N, CuratorMAS reaches 95.55\% recall, a 7.73-point improvement over the prompt-based method, 
while maintaining a precision of 73.69\% and reducing the residual noise rate to 3.64\%. Compared with 
model-based baselines, CuratorMAS drastically reduces the noise rate with consistently higher recall 
and precision. Although LLMs are effective judges (71.72\% precision and 87.82\% recall), CuratorMAS decomposes 
the complex curation procedure into a programmable workflow, thereby further improving its overall performance. 
Representative examples are shown in Figure~\ref{fig:cifar_noise_examples}.
On NoisyAG-News, CuratorMAS also demonstrates its robust capabilities with the lowest residual noise rate of 10.60\% and
the highest precision and recall. 

These results underscore CuratorMAS's consistent ability to foster more effective
collaboration that translates directly into curation capability, regardless of the task domain and modality. 
We further observe a precision--recall trade-off: additional filtering rounds improve noise coverage 
but may introduce a higher FPR, indicating the need to balance automated filtering with the available human-review budget.

\begin{figure*}[t]
  \centering
  \includegraphics[width=0.87\linewidth]{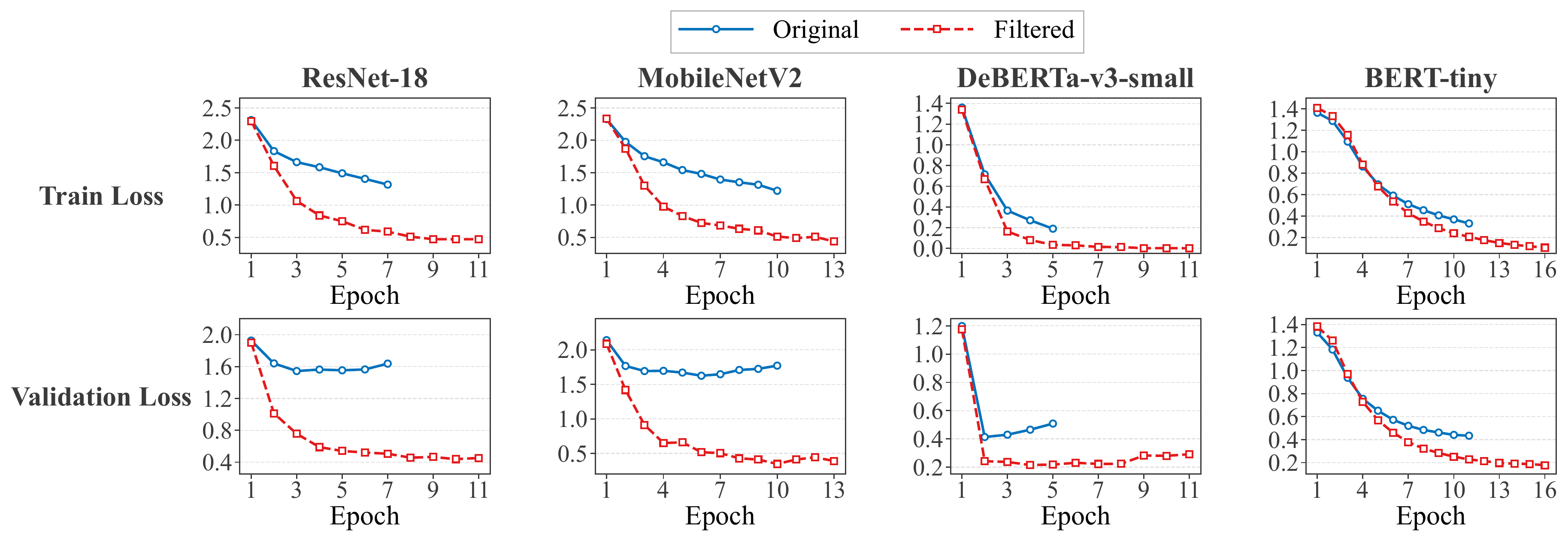}
  \caption{Training and validation loss curves of downstream models trained on the original and CuratorMAS-curated datasets.}
  \label{fig:downstream_loss}
\end{figure*}

\subsection{Impact on Downstream Training}
\label{sec:downstream_training}
To further investigate how CuratorMAS benefits model training and generalization, 
we evaluate the downstream performance of models trained on the aforementioned CIFAR-10N and 
NoisyAG-News datasets curated by CuratorMAS. Figure~\ref{fig:downstream_performance} shows the overall 
performance gains across eight model architectures. For instance, on CIFAR-10N, all four models consistently 
achieve performance gains, with F1 scores improving by up to 8.88 pp (MobileNetV2) and by 5.99 pp on average. 
Similarly, on NoisyAG-News, all four BERT-style models benefit from the 
curated dataset with an average F1 score improvement of 3.64 pp. These results indicate that CuratorMAS 
supports better downstream model performance across both text and image modalities.

Beyond improving downstream performance, CuratorMAS also yields more favorable training dynamics. As 
illustrated in Figure~\ref{fig:downstream_loss}, models trained on the curated datasets converge in 
fewer epochs and reach lower final training losses than those trained on the original noisy 
datasets. The validation curves also exhibit smaller fluctuations and less evidence of overfitting, 
suggesting more stable optimization. Together with the improvements in test-set performance, these 
observations indicate that CuratorMAS improves optimization efficiency and downstream generalization.

\subsection{Ablation Study}

\input{tables/backbone_real_world_comparison.tex}

\paragraph{Sensitivity Analysis.}
We conduct a sensitivity analysis using three different LLM backbones for our task-execution agents. Table~\ref{tab:filtering_backbone_family_overview} 
reveals noticeable performance differences among the three models. GPT-5.4 delivers the best overall performance, 
achieving a favorable trade-off among precision, recall, and residual noise rate. 
GPT-5.3-Codex does not outperform either GPT-5.4 or Sonnet-4.6. This result suggests 
that specialization for coding alone does not necessarily translate into stronger data-curation capabilities, 
whereas general reasoning ability may play a more fundamental role in data curation.
This observation is broadly consistent with previous research~\cite{LLMsStrugglewithDataAnalysis}. 
These findings indicate that although CuratorMAS can effectively support automated data curation, 
its curation performance remains sensitive to the capabilities of the underlying LLM backbone.

\paragraph{Component Ablation.}
We also analyze the contributions of two key components of CuratorMAS on NoisyAG-News: external knowledge retrieval and 
skill enhancement. To prevent data leakage and ensure a fair comparison, the domain skills are constructed using a held-out subset.
As shown in Table~\ref{tab:ablation_noisyagnews}, both components  
contribute to the effectiveness of CuratorMAS. Removing both leads to the largest 
degradation, resulting in the poorest curation performance. External knowledge retrieval supports domain-specific 
evaluation, whereas domain skills facilitate long-term knowledge accumulation and stability.
These results demonstrate the complementary roles of the two components in CuratorMAS.

\input{tables/ablation_table.tex}

\section{Conclusion}
We presented CuratorMAS, a multi-agent collaboration framework that enables agents to flexibly evaluate and 
curate datasets across diverse domains. By breaking down the complex data curation process into executable and 
programmable stages, CuratorMAS can adapt its evaluation criteria to different domains 
and dynamic user queries, thereby providing a structured paradigm for automated data curation. Experiments on three 
datasets against six baselines demonstrate the effectiveness of CuratorMAS, which reduces the noise rate by up to 36.03 pp. 
Moreover, models trained on the curated datasets achieve improvements of up to 8.9 pp in F1 score. 
These results establish CuratorMAS as a practical and flexible solution to automated data curation across diverse real-world scenarios.
Future work includes extending CuratorMAS to more complex tasks while further improving its generalization and stability 
for high-stakes applications.

\bibliography{aaai2027}

\end{document}

%% file: algorithm.tex
\begin{algorithm}[tb]
\caption{End-to-end workflow of CuratorMAS}
\label{alg:autocuration}
\textbf{Input}: Dataset $D$, query $q$, skill library $\mathcal{S}$, tool library $\mathcal{T}$\\
\textbf{Output}: Curated dataset $D_{\mathrm{curated}}$, rule-wise noise sets
$D_{\mathrm{noise}}^{(\ell)}$, updated $S'_{\mathrm{domain}}$ and $T'_{\mathrm{domain}}$
\begin{algorithmic}[1]

\STATE \textcolor{blue}{$\triangleright$ \textbf{Dataset Comprehension}}
\STATE $(\mathcal{U}_{D}, \mathcal{C}_{q})
       \sim \pi_{\mathrm{explorer}}(\cdot \mid D,q)$
\STATE $S_{\mathrm{domain}}
       \gets \mathrm{LoadSkill}(\mathcal{S},q)$
\STATE $T_{\mathrm{domain}}
       \gets \mathrm{LoadTool}(\mathcal{T},q)$

\STATE \textcolor{blue}{$\triangleright$ \textbf{External Knowledge Retrieval}}
\STATE $\{x_i\}_{i=1}^{N}
       \sim \pi_{\mathrm{researcher}}(\cdot \mid S_{\mathrm{domain}})$
\FOR{$i=1$ \TO $N$}
    \STATE $\{d_{i,j}\}_{j=1}^{K}
           \gets \mathrm{TopK}(\mathrm{Search}(x_i),K)$
    \STATE $\{s_{i,j}\}_{j=1}^{K}
           \sim \pi_{\mathrm{subagent}}
           (\cdot \mid \{d_{i,j}\}_{j=1}^{K})$
\ENDFOR
\STATE $Q_{\mathrm{comp}}
       \sim \pi_{\mathrm{researcher}}
       (\cdot \mid \mathcal{U}_{D},\{s_{i,j}\},S_{\mathrm{domain}})$

\STATE \textcolor{blue}{$\triangleright$ \textbf{Knowledge-Augmented Quality Evaluation}}
\STATE $\mathcal{K}_{\mathrm{aug}}
       \gets Q_{\mathrm{comp}} \oplus S_{\mathrm{domain}}$
\STATE $P_{\mathrm{eval}}
       \sim \pi_{\mathrm{evaluator}}
       (\cdot \mid \mathcal{U}_{D},\mathcal{K}_{\mathrm{aug}},\mathcal{C}_{q})$
\FORALL{$t_m \in P_{\mathrm{eval}}$ in parallel}
    \STATE $o_m
           \sim \pi_{\mathrm{worker}}(\cdot \mid t_m,\mathcal{U}_{D})$
\ENDFOR
\STATE $\mathcal{B}
       \gets \mathrm{Buffer}(\{o_m\}_{m=1}^{M})$

\STATE \textcolor{blue}{$\triangleright$ \textbf{Agentic Filtering and Dataset Export}}
\STATE $\{r_{\ell}\}_{\ell=1}^{L}
       \sim \pi_{\mathrm{filter}}
       (\cdot \mid \mathcal{B},\mathcal{U}_{D},\mathcal{C}_{q})$
\FOR{$\ell=1$ \TO $L$}
    \STATE $D_{\mathrm{noise}}^{(\ell)}
           = \{x \in D : r_{\ell}(x)=1\}$
\ENDFOR
\STATE $D_{\mathrm{noise}}
       = \displaystyle\bigcup_{\ell=1}^{L}
       D_{\mathrm{noise}}^{(\ell)}$
\STATE $D_{\mathrm{curated}}
       = D \setminus D_{\mathrm{noise}}$

\STATE \textcolor{blue}{$\triangleright$ \textbf{Experience-Based Self-Evolution}}
\STATE $S_{\mathrm{patch}}
       \sim \pi_{\mathrm{refiner}}
       (\cdot \mid P_{\mathrm{eval}},\mathcal{B},S_{\mathrm{domain}})$
\STATE $S'_{\mathrm{domain}}
       \gets S_{\mathrm{domain}} \cup S_{\mathrm{patch}}$
\STATE $T_{\mathrm{patch}}
       \sim \pi_{\mathrm{refiner}}
       (\cdot \mid \mathcal{B},T_{\mathrm{domain}})$
\STATE $T'_{\mathrm{domain}}
       \gets T_{\mathrm{domain}}
       \cup \mathbb{I}_{\mathrm{gate}}(T_{\mathrm{patch}})$

\RETURN $D_{\mathrm{curated}}, D_{\mathrm{noise}}^{(\ell)},
        S'_{\mathrm{domain}}, T'_{\mathrm{domain}}$
\end{algorithmic}
\end{algorithm}

%% file: tables/contamination_comparision.tex
\begin{table}[t]
  \centering
  \small
  \setlength{\tabcolsep}{3pt}
  \begin{tabular*}{\linewidth}{@{\extracolsep{\fill}}lrrr}
    \toprule
    \textbf{Metric} & \textbf{10\%} & \textbf{20\%} & \textbf{30\%} \\
    \midrule
    Precision & 76.97 $\pm$ 9.70 & \underline{78.13} $\pm$ 28.61 & \textbf{92.72} $\pm$ 7.06 \\
    Recall & \underline{80.00} $\pm$ 9.54 & 78.83 $\pm$ 16.07 & \textbf{83.00} $\pm$ 8.97 \\
    F1 & \underline{78.22} $\pm$ 8.08 & 77.96 $\pm$ 22.56 & \textbf{87.13} $\pm$ 2.51 \\
    FPR & \textbf{2.80} $\pm$ 1.44 & 7.32 $\pm$ 10.25 & \underline{3.44} $\pm$ 3.84 \\
    \bottomrule
  \end{tabular*}
  \caption{Performance of CuratorMAS across contamination levels.
  Metrics (\%) are reported as mean $\pm$ std across three runs,
  with contaminated samples as the positive class.}
  \label{tab:controlled_contamination_main}
\end{table}

%% file: tables/baseline_real_world_comparison.tex
\begin{table*}[t]
  \centering
  \small
  \setlength{\tabcolsep}{3.5pt}
  \begin{tabular*}{0.9\textwidth}{@{\extracolsep{\fill}}llrrrrrr}
    \toprule
    \textbf{Dataset} & \textbf{Method} & \textbf{Filter} & \textbf{Hit} & \textbf{Precision} & \textbf{Recall} & \textbf{F1} & \textbf{Noise Level} \\
    \midrule

    \multirow{12}{*}{\textbf{CIFAR-10N}}
      & Original
        & - & - & - & - & - & 39.67\% \\

    \cmidrule(l){2-8}
      & \multicolumn{7}{l}{\cellcolor{gray!15}\textit{Rule-based Method(s)}} \\
      & Heuristic
        & 1,188 & 556 & 46.80\% & 46.72\% & 46.76\% & 34.99\% \\

    \cmidrule(l){2-8}
      & \multicolumn{7}{l}{\cellcolor{gray!15}\textit{Model-based Method(s)}} \\
      & CleanLab
        & 876 & 727 & \textbf{82.99\%} & 61.09\% & 70.38\% & 21.80\% \\

      & DivideMix
        & 1,539 & 927 & 60.23\% & 77.90\% & 67.94\% & 18.00\% \\

      & ProMix
        & 1,180 & 832 & 70.51\% & 69.92\% & 70.21\% & 19.67\% \\

      & LossRank
        & 1,188 & 779 & 65.57\% & 65.46\% & 65.52\% & 22.68\% \\

    \cmidrule(l){2-8}
      & \multicolumn{7}{l}{\cellcolor{gray!15}\textit{Prompt-based Method(s)}} \\
      & LLM-as-a-Judge
        & 1,457 & 1045 & 71.72\% & \underline{87.82\%} & \underline{78.96\%} & \underline{9.40\%} \\

    \cmidrule(l){2-8}
      & \multicolumn{7}{l}{\cellcolor{gray!15}\textit{Agentic Method(s)}} \\
      & \textbf{Ours}
        & 1,543 & 1137 & \underline{73.69\%} & \textbf{95.55\%} & \textbf{83.21\%} & \textbf{3.64\%} \\

    \midrule

    \multirow{13}{*}{\textbf{NoisyAG-News}}
      & Original
        & - & - & - & - & - & 19.83\% \\

    \cmidrule(l){2-8}
      & \multicolumn{7}{l}{\cellcolor{gray!15}\textit{Rule-based Method(s)}} \\
      & Heuristic
        & 507 & 202 & 39.84\% & 33.95\% & 36.66\% & 15.76\% \\

    \cmidrule(l){2-8}
      & \multicolumn{7}{l}{\cellcolor{gray!15}\textit{Model-based Method(s)}} \\
      & CleanLab
        & 99 & 88 & \textbf{88.89\%} & 14.79\% & 25.36\% & 17.48\% \\

      & DivideMix
        & 133 & 108 & \underline{81.20\%} & 18.15\% & 29.67\% & 16.99\% \\

      & ProMix
        & 665 & 275 & 41.35\% & 46.22\% & 43.65\% & 13.70\% \\

      & LossRank
        & 594 & 324 & 54.55\% & \textbf{54.45\%} & \underline{54.50\%} & \underline{11.26\%} \\

    \cmidrule(l){2-8}
      & \multicolumn{7}{l}{\cellcolor{gray!15}\textit{Prompt-based Method(s)}} \\
      & LLM-as-a-Judge
        & 240 & 177 & 73.75\% & 29.75\% & 42.40\% & 15.14\% \\

    \cmidrule(l){2-8}
      & \multicolumn{7}{l}{\cellcolor{gray!15}\textit{Agentic Method(s)}} \\
      & \textbf{Ours}
        & 425 & 322 & 75.76\% & \underline{54.12\%} & \textbf{63.14\%} & \textbf{10.60\%} \\

    \bottomrule
  \end{tabular*}
  \caption{Overall performance comparison with baselines on CIFAR-10N and NoisyAG-News.}
  \label{tab:filtering_baseline_comparison}
\end{table*}

%% file: tables/backbone_real_world_comparison.tex
\begin{table}[t]
  \centering
  \small
  \setlength{\tabcolsep}{2.5pt}
  \begin{tabular*}{\columnwidth}{@{\extracolsep{\fill}}llrrr}
    \toprule
    \textbf{Dataset} & \textbf{Method} & \textbf{Precision} & \textbf{Recall} & \textbf{Noise} \\
    \midrule

    \multirow{5}{*}{\textbf{CIFAR-10N}}
      & \multicolumn{4}{l}{\cellcolor{gray!15}\textit{GPT-family}} \\
      & GPT-5.4
        & \textbf{73.69\%} & \textbf{95.55\%} & \textbf{3.64\%} \\
      & GPT-5.3-Codex
        & 60.76\% & 94.20\% & 5.97\% \\

    \cmidrule(l){2-5}
      & \multicolumn{4}{l}{\cellcolor{gray!15}\textit{Claude-family}} \\
      & Sonnet-4.6
        & \underline{61.04\%} & \underline{95.46\%} & \underline{4.74\%} \\

    \midrule

    \multirow{5}{*}{\textbf{NoisyAG-News}}
      & \multicolumn{4}{l}{\cellcolor{gray!15}\textit{GPT-family}} \\
      & GPT-5.4
        & \textbf{75.76\%} & 54.12\% & 10.60\% \\
      & GPT-5.3-Codex
        & 54.55\% & \underline{63.53\%} & \underline{9.41\%} \\

    \cmidrule(l){2-5}
      & \multicolumn{4}{l}{\cellcolor{gray!15}\textit{Claude-family}} \\
      & Sonnet-4.6
        & \underline{58.76\%} & \textbf{65.38\%} & \textbf{8.81\%} \\

    \bottomrule
  \end{tabular*}
  \caption{Sensitivity of CuratorMAS to LLM Backbones on CIFAR-10N and NoisyAG-News.}
  \label{tab:filtering_backbone_family_overview}
\end{table}

%% file: tables/ablation_table.tex
\begin{table}[t]
  \centering
  \begin{tabular*}{0.95\linewidth}{@{\extracolsep{\fill}}lcccc}
    \toprule
    \textbf{Setting} & \textbf{Filtered} & \textbf{Precision} & \textbf{Recall} & \textbf{Noise} \\
    \midrule
    \textit{\textbf{w/o KA}}          & 357 & \underline{65.27\%} & 39.16\% & 13.70\% \\
    \textit{\textbf{w/o Skill}}       & 425 & \textbf{75.76\%} & \underline{54.12\%} & \underline{10.60\%} \\
    \textit{\textbf{Neither}} & 433 & 29.56\% & 21.51\% & 18.19\% \\
    \textit{\textbf{Full}}            & 643 & 57.85\% & \textbf{62.52\%} & \textbf{9.46\%} \\
    \bottomrule
  \end{tabular*}
  \caption{Component Ablation of CuratorMAS.
  \textit{w/o KA} uses only domain skills;
  \textit{w/o Skill} uses only external knowledge retrieval;
  \textit{Neither} removes both components;
  \textit{Full} denotes the complete CuratorMAS. }
  \label{tab:ablation_noisyagnews}
\end{table}